\documentclass{egpubl}
\usepackage{pg2025s}
\usepackage{amsmath}
\usepackage{amssymb}
\usepackage{multirow}
\usepackage{microtype}
\usepackage{wrapfig}

\usepackage{xcolor}

\newcommand{\norm}[1]{\lVert#1\rVert}

\WsConferencePaper

\usepackage[T1]{fontenc}
\usepackage{dfadobe}  

\usepackage{cite}  
\BibtexOrBiblatex
\electronicVersion
\PrintedOrElectronic
\ifpdf \usepackage[pdftex]{graphicx} \pdfcompresslevel=9
\else \usepackage[dvips]{graphicx} \fi

\usepackage{egweblnk} 

\title[Neural Shadow Art]%
      {Neural Shadow Art}

\author[C. Wang et al.]
{\parbox{\textwidth}{\centering Caoliwen Wang$^{1}$\orcid{0009-0004-3822-9094},
        Bailin Deng$^{2}$\orcid{0000-0002-0158-7670} and Juyong Zhang\thanks{Corresponding author: juyong@ustc.edu.cn}$^{1}$\orcid{0000-0002-1805-1426}
        }
        \\
            $^{1}$University of Science and Technology of China, China 
    \qquad
    $^{2}$Cardiff University, United Kingdom
}

\begin{document}

\teaser{
 \centering
\includegraphics[width=\linewidth]{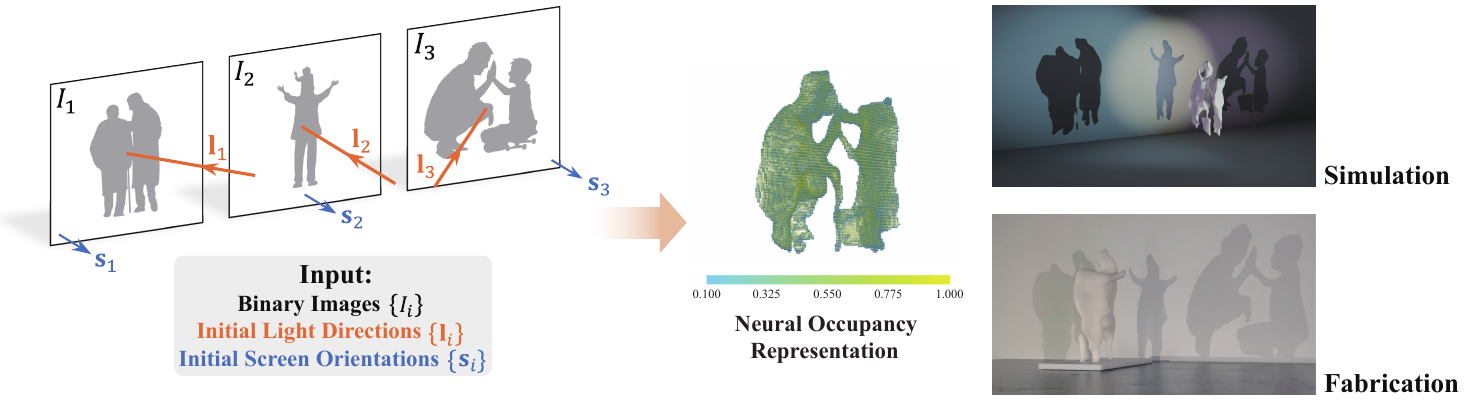}
  \caption{We introduce Neural Shadow Art, providing a novel neural implicit occupancy representation for shadow art, which significantly expands the possibilities of shadow art and makes this art form even more awe-inspiring.}
\label{fig:teaser}
}

\maketitle
\begin{abstract}
Shadow art is a captivating form of sculptural expression where the projection of a sculpture in a specific direction reveals a desired shape with high precision. In this work, we introduce Neural Shadow Art, which leverages implicit occupancy function representation to significantly expand the possibilities of shadow art. This representation enables the design of high-quality, 3D-printable geometric models with arbitrary topologies at any resolution, surpassing previous voxel- and mesh-based methods. Our method provides a more flexible framework, enabling projections to match input binary images under various light directions and screen orientations, without requiring light sources to be perpendicular to the screens. Furthermore, we allow rigid transformations of the projected geometries relative to the input binary images and simultaneously optimize light directions and screen orientations to ensure that the projections closely resemble the target images, especially when dealing with inputs of complex topologies. In addition, our model promotes surface smoothness and reduces material usage. This is particularly advantageous for efficient industrial production and enhanced artistic effect by generating compelling shadow art that avoids trivial, intersecting cylindrical structures. In summary, we propose a more flexible representation for shadow art, significantly improving projection accuracy while simultaneously meeting industrial requirements and delivering awe-inspiring artistic effects.


\end{abstract}

\section{Introduction}

\begin{figure}[t]
\centering
\includegraphics[width=1.\columnwidth]{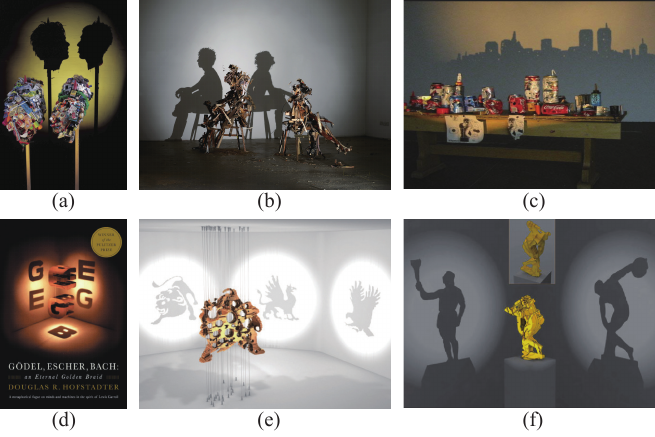}
\caption{\textbf{Various Shadow Art Examples.} (a)-(c): Examples of historically renowned shadow art created by artists Tim Noble and Sue Webster, where carefully positioning materials produces artistic effects in the projection from a specific direction. (d)-(f): Shadow art pieces—including one from the cover of Hofstadter’s ``Gödel, Escher, Bach'' and others adapted from the works of Mitra and Pauly~\cite{MP09}, and Schwartzburg and Pauly~\cite{SP12}—demonstrate how varying light conditions and screen orientations can reveal distinct artistic content, aligning with the computer graphics community’s conception of shadow art.
} 
\label{fig:Intro}
\end{figure}

Shadow art is a unique form of artistic expression that employs shadows as media for information transmission. Artists have long demonstrated remarkable skills in creating intricate geometric structures and light fields that project desired images, utilizing spatial projections of objects to convey artistic visions. Historically, shadow art has manifested in numerous captivating exhibitions. For instance, Tim Noble and Sue Webster assemble everyday objects into seemingly random three-dimensional configurations that, when illuminated from specific directions, cast highly detailed and lifelike shadows, such as a human face profile (Fig.~\ref{fig:Intro}(a)) or two people sitting back-to-back (Fig.~\ref{fig:Intro}(b)). Fig.~\ref{fig:Intro}(c) shows their simulation of a cityscape at sunrise, where a warm-toned light field creates a recognizable scene. These artist-created works depend on the precise spatial arrangement of materials to produce visually striking patterns on a screen when illuminated from a specific direction. This creative process demands exceptional spatial reasoning, particularly for constructing geometries that yield different projections under varying light conditions, as seen in various examples (Fig.~\ref{fig:Intro}(d-f)). Fig.~\ref{fig:Intro}(d) displays the cover of Douglas Hofstadter's "Gödel, Escher, Bach," featuring geometric forms that cast shadows of the letters G, E, and B when viewed from three orthogonal directions. Another example is the pioneering computational shadow art from~\cite{MP09}  (Fig.~\ref{fig:Intro}(e)), which projects a complex geometry along three non-orthogonal directions. Finally, Fig.~\ref{fig:Intro}(f) is derived from \cite{SP12}, where a sculpture projects two distinct silhouettes—a figure preparing to throw an object and another holding a torch—showcasing remarkable artistic expression through careful shadow manipulation.

These examples of shadow art are both engaging and thought-provoking, exemplifying the precise calculation and control of light and three-dimensional geometry. As achieving the intended artistic effect through intuition and spatial imagination alone becomes exceedingly challenging especially for non-professionals, the development of generalizable algorithms for designing shadow art is essential. In computer graphics, shadow art creation is framed as an inverse problem to shadow generation: instead of generating shadows from a known scene, algorithms take desired shadow images as input to compute a three-dimensional geometry that will cast the intended shadows. This task requires sophisticated geometric optimization and manipulation to yield results that are both aesthetically compelling and technically accurate.

Many studies have been dedicated to the design of shadow art. Mitra and Pauly~\cite{MP09} were the first to introduce shadow art to the computer graphics community, formally addressing the problem through a geometric optimization method using explicit voxel representation. However, this method allows for as rigid as possible deformations of the input image, which can lead to notable discrepancies between the final projections and the input figures. Sadekar et al.~\cite{STR22} proposed a shadow art design method that better aligns with the input by utilizing voxel- and mesh-based differentiable rendering. However, this method has high requirements for the input images and cannot produce satisfactory results when faced with conflicting shadow constraints. Additionally, the voxel-based approach is limited by the given resolution, while the mesh-based approach is constrained by the initial topology, often leading to significant flipping and resulting in poor geometric properties unsuitable for production. Both of the above methods overlook the optimization of complex light conditions, which significantly restricts flexibility. The optimization of volume is also neglected; as a result, some of the outcomes resemble trivial intersections of cylindrical shapes, ultimately diminishing the artistic visual appeal that is central to shadow art.

To address these limitations, we propose Neural Shadow Art, which provides a novel and flexible occupancy representation of shadow art. In this formulation, we are no longer constrained by a predefined topology and can recover high-quality, 3D-printable smooth geometry at arbitrary resolutions, surpassing previous voxel- and mesh-based methods. In our framework, we no longer require the light direction to be perpendicular to the projection plane. This expands the possibilities of shadow art, making the resulting artistic effects more vivid. In addition to using differentiable rendering techniques to approximate the input images, we allow for rigid deformation of the input images while also enabling a joint optimization of light conditions. This significantly improves the accuracy of the projection results, particularly for input images with complex topologies, better meeting users' demands for shadow effects. Furthermore, we optimize the volume of the geometry, which not only better satisfies the industrial requirement of minimizing material usage but also aligns with the goal of shadow art as an art form, achieving the desired shadow effect with less material while maintaining high-quality geometric properties that are suitable for 3D printing. In contrast to trivial solutions resembling cylindrical intersections, our results often produce more interesting artistic effects. In summary, our main contributions include: 
\begin{itemize}
    \item We present a novel neural implicit representation of shadow art, offering greater flexibility. This representation removes the constraints of predefined topology and allows for high-quality, 3D-printable smooth geometry at any resolution, surpassing previous voxel- and mesh-based methods.
    \item Unlike earlier works, our approach allows the light direction to be non-perpendicular to the projection plane, offering more possibilities for model generation. It also supports rigid transformations of the input image and joint optimization of the light direction and screen orientation, ensuring accurate projections.
    \item We incorporate geometric smoothness and volume optimization into our model, which not only meets industrial production requirements but also highlights the intricacy and elegance of this art form, avoiding trivial solutions such as intersections of cylindrical shapes.
\end{itemize}
\section{Related Works}
\paragraph*{Shadow Information.} Many works have analyzed the information that can be extracted from shadows and applied it to tasks such as shape reconstruction or generation. Waltz~\cite{Wal03} pioneered the use of shadows to infer 3D shapes in line drawings. Building on this work, Steven and Kanade~\cite{SK83} developed a method utilizing shadow collections to infer surface orientations for both polyhedra and curved surfaces. Bouguet and Perona~\cite{BJ99}  introduced a "weak structured lighting" system that extracts the 3D shapes of objects by observing the spatio-temporal locations of cast shadows.  Savarese et al.~\cite{SM07} proposed a reconstruction system using silhouettes and shadow carving, demonstrating how shadows reveal unique information, such as object concavities. More recently, deep learning techniques have been employed to learn information from shadows for reconstruction or generation tasks. For instance, Kushagra et al.~\cite{TKR22} explored learning implicit scene representations from shadow information via volume rendering, while Liu et al.~\cite{LMM*23} used generative models to predict 3D object geometry from shadows, achieving promising results, especially for partially or fully occluded objects. These advancements underscore the evolving role of shadows in extracting and refining 3D structural information within computer graphics and vision.
 
\paragraph*{Shadow Art.} Various works have addressed the computational generation of shadow art. The problem was formally introduced to the computer graphics community by Mitra and Pauly~\cite{MP09}, who proposed an explicit voxel-based optimization method for constructing 3D geometries that cast specific projections. A limitation of their approach is the As-Rigid-As-Possible (ARAP) deformation applied during optimization, which can lead to excessive distortion in the final projection compared to the input image. Sadekar et al.~\cite{STR22} later introduced a differentiable rendering framework for the same objective, demonstrating its applicability to other artistic forms like geometry prediction from half-toned images. However, their method may produce poor results if input images have inconsistent configurations, as it does not process them directly. Additionally, their voxel-based approach is resolution-limited, and their mesh-based approach is constrained by its initial topology.

Beyond these direct methods for generating shadow art effects similar to ours, other artistic expressions also utilize shadows. Won et al.~\cite{Shadowtheater} explored extracting motion sequences from shadow shapes. Min et al.~\cite{MLWL17} presented an algorithm for generating grayscale shadows using area lights or arrays of point lights with occluding objects, thereby expanding the scope of light sources used. Hsiao et al.~\cite{HHC18} developed an algorithm for creating multi-view wire sculptures whose projections form specified line drawings. Qu et al.~\cite{qu2023dreamwire} introduced semantically-guided wire art design, and Tojo et al.~\cite{KA24} developed a workflow for fabricating wire art from various inputs, including a 3D-printable jig structure to produce the generated wire paths. Gangopadhyay et al.~\cite{handshadowart1} and Xu et al.~\cite{handshadow} were dedicated to generating shadows by leveraging diverse hand poses. Further artistic works are discussed in Wu et al.'s review on 3D visual optical art design \cite{WFCL22}. 

\paragraph*{Representations for 3D Learning.} Learning-based 3D reconstruction techniques are often categorized by their underlying data representations: voxel-based, point cloud-based, mesh-based, or implicit function-based.
Voxel-based methods \cite{WSK*15,MS15,BLRW16,SGF16} 
directly extend 2D pixel grids but are constrained by resolution, especially with sparse data, and suffer from high storage requirements as resolution increases.
Point cloud-based methods \cite{RQSMG16,QYSG17,KL17,YFST18} are more memory-efficient but lack explicit topological information, often necessitating significant post-processing. 
Mesh-based methods can represent topology and avoid post-processing, but learning techniques do not readily extend to their irregular structure, and such data can suffer from noise, missing data, and resolution issues, as noted in works like~\cite{CRB*16}.

Implicit representations of geometry, however, circumvent many of these issues and are increasingly prevalent in 3D learning research. Park et al.~\cite{PFS*19} proposed learning Signed Distance Functions (SDFs) to represent 3D geometry, significantly reducing storage while preserving complex topologies and showing potential for shape completion tasks. 
Mescheder et al.~\cite{MON*19} used occupancy networks, extending binary occupancy to continuous values in $[0,1]$, demonstrating strong performance in both supervised and unsupervised learning scenarios. These two approaches represent common forms of implicit geometric representations in current 3D learning.  
Building on implicit functions, Mildenhall et al.~\cite{MST*21} introduced Neural Radiance Fields (NeRF), which represent scenes with a neural implicit function guided by neural rendering.  
Wang et al.~\cite{WLL*21} trained SDF representations for object reconstruction under neural rendering guidance, and Long et al.~\cite{LLL*23} extended similar ideas to unsigned distance functions, enabling reconstruction of non-closed surfaces. While SDFs are widely used, the precise measurement of geometric volume using them remains an open area to our knowledge. Our work shares similarities with these studies in its use of neural implicit representations for geometric modeling.

\begin{figure*}[!htb]
\centering
\includegraphics[width=1.0\textwidth]{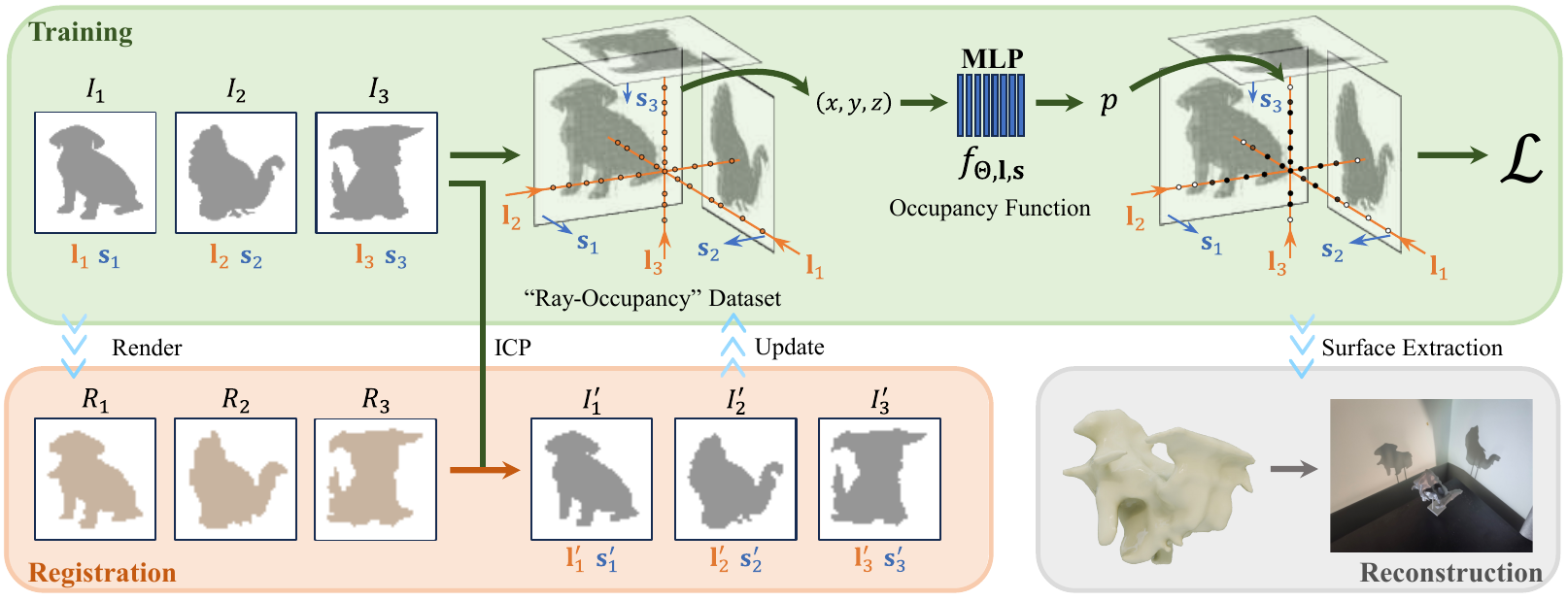}
\caption{\textbf{Neural Shadow Art Pipeline.} At each epoch, a ``Ray-Occupancy'' dataset is generated from current inputs to supervise the neural implicit occupancy function via backpropagation. Light directions $\{\mathbf{l}_i\}$ and screen orientations $\{\mathbf{s}_i\}$ are jointly optimized. To accurately account for discrepancies between shadows and their intended targets, we periodically render shadows $\{R_i\}$ and register them with input images $\{I_i\}$ to update training targets $\{I'_i\}$ for subsequent training. Finally, the 3D surface is extracted from the learned implicit occupancy representation for simulation and manufacturing.} \label{fig:pipeline}
\end{figure*}
\section{Method}\label{sec-method}
We take as input a set of binary images $\{I_i\}$, along with their initial light directions $\{\mathbf{l}_i 
\in \mathbb{R}^3\}$ where $\mathbf{l}_i$ is a unit vector, and their corresponding screen orientations $\{\mathbf{s}_i \in \mathbb{R}^3\}$ where $\mathbf{s}_i$ is the unit normal of the projection screen and points towards the object. By definition, \(\langle \mathbf{l}_i, \mathbf{s}_i \rangle<0\). Our primary objective is to generate a 3D geometry and simultaneously optimize the light directions  $\{\mathbf{l}_i\}$ and screen orientations $\{\mathbf{s}_i\}$ such that the cast shadows on each screen closely match the target images $\{I_i\}$. 

Fig.~\ref{fig:pipeline} presents an overview of our pipeline. The core of our method is an iterative process.  At each epoch, we first generate a ``Ray-Occupancy'' dataset based on the current inputs.
Using this dataset, we compute a single loss function and use backpropagation to simultaneously update all learnable parameters: the weights of the neural implicit occupancy function that represents the 3D geometry, the light directions $\{\mathbf{l}_i\}$, and the screen orientations $\{\mathbf{s}_i\}$. To accurately account for the differences between the shadows and their targets, we periodically render the current shadow projections $\{R_i\}$ and perform registration between these shadows and the original input images $\{I_i\}$ to update the target images $\{I'_i\}$. 
These updated images are then used for training in subsequent epochs. 
Finally, once training converges, the 3D surface is extracted from the learned neural implicit occupancy representation for simulation and manufacturing. 

\subsection{Occupancy Function} \label{subsec:occu_net}
At the core of our method is a neural occupancy function \( f_{\Theta,\mathbf{l},\mathbf{s}}: \mathbb{R}^3 \mapsto [0, 1] \) that maps a 3D coordinate to an occupancy probability:
\begin{equation}
f_{\Theta,\mathbf{l},\mathbf{s}}(x, y, z) = p\in [0,1], \label{equa:occu_def}
\end{equation}
where \(\Theta\) denotes the parameters of the neural network, \(\mathbf{l} =\{\mathbf{l}_i\}\) is the set of current light directions, and \(\mathbf{s} =\{\mathbf{s}_i\}\) is the set of current screen orientations. The input $(x,y,z)$  is a 3D coordinate within a normalized space, and the output $p$ is the predicted occupancy probability at that point. A value of $p=0$ indicates that the point is not occupied, while values closer to 1 indicate a higher likelihood that the point is occupied. During surface reconstruction, points with an occupancy value greater than a threshold \(\tau\) (typically 0.5 in our framework) are considered to be part of the solid geometry. Unlike some occupancy network definitions such as~\cite{MON*19}, the input of our function consists solely of 3D coordinates, without explicit view-dependent information.

We utilize an MLP network to implement the occupancy function \( f_{\Theta, \mathbf{l}, \mathbf{s}} \). For any input 3D coordinate \(\mathbf{p}\), we first apply positional encoding \( \gamma(\mathbf{p}) \) to map the input coordinates to a higher-dimensional feature space:
\begin{equation}
    \gamma(\mathbf{p}) = \left( \mathbf{p}, \sin(2^0  \mathbf{p}), \cos(2^0  \mathbf{p}),  \dots, \sin(2^{L-1}  \mathbf{p}), \cos(2^{L-1}  \mathbf{p}) \right),
\label{equa:gamma_p}
\end{equation}
where \(L\) is a hyperparameter determining the dimensionality of the encoding.
The positionally encoded features then pass through an MLP \( \widetilde{f}_{\Theta, \mathbf{l}, \mathbf{s}} \) to obtain the predicted occupancy value of $\mathbf{p}$:
\begin{equation}
    f_{\Theta,\mathbf{l},\mathbf{s}}(\mathbf{p}) = \widetilde{f}_{\Theta,\mathbf{l},\mathbf{s}} (\gamma(\mathbf{p})).
    \label{eq:MLPFunc}
\end{equation}

\subsection{Training}\label{one_epoch_training}

Training the neural occupancy function $f_{\Theta,\mathbf{l},\mathbf{s}}$ involves iteratively refining its parameters $\Theta$ along with the light directions $\{\mathbf{l}_i\}$ and screen orientations $\{\mathbf{s}_i\}$. This process relies on a specially constructed dataset and a composite loss function.

\paragraph*{``Ray-Occupancy'' Dataset.} At the beginning of each training epoch, we construct a ``Ray-Occupancy'' dataset. 
This dataset includes a large number of rays cast through the 3D scene towards the target shadow images, along with their corresponding ground-truth occupancy labels indicating whether a ray should be occluded or unoccluded by the geometry based on the target image pixels.
It serves as the primary input for the subsequent loss computation and network training steps.

\begin{figure}[tb]
\centering
\includegraphics[width=0.6\columnwidth]{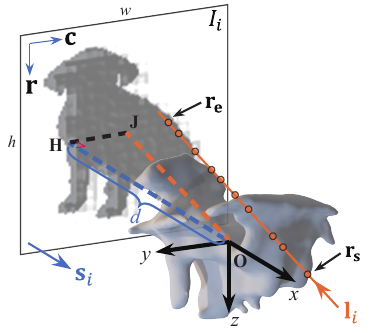}
\caption{\textbf{Visualization of the Quantities in Eqs. (\ref{equa:r_s})-(\ref{equa:r}) for the ``Ray-Occupancy'' Dataset.}} 
\label{fig:Ray-Occupancy Visual}
\end{figure}

For each target image $I_i$ with its corresponding light direction $\mathbf{l}_i$ and screen orientation $\mathbf{s}_i$, we define rays that pass through the projection screen.
As illustrated in Fig.~\ref{fig:Ray-Occupancy Visual}, let \( \mathbf{O} \) be the origin of the normalized 3D space, and $\mathbf{J}$ be the center of $I_i$ on the screen. We require the vector $\overrightarrow{\mathbf{OJ}}$ to be parallel to $\mathbf{l}_i$. Moreover, let $\mathbf{H}$ be the projection of $\mathbf{O}$ onto the screen, so that $\overrightarrow{\mathbf{OH}}$ is parallel to $\mathbf{s}_i$.
The ray corresponding to a pixel $(p_x, p_y)$ in the target image on the screen is defined by its start point $\mathbf{r_s}$ and end point $\mathbf{r_e}$. They are calculated as:
\begin{align}
      \mathbf{r_e} & = \norm{\overrightarrow{\mathbf{OJ}}}\mathbf{l}_i+ \frac{w}{h}(\frac{p_x}{w}-\frac{1}{2})\mathbf{c} +(\frac{p_y}{h}-\frac{1}{2})\mathbf{r},
    \label{equa:r_s}  \\
    \mathbf{r_s} &= \mathbf{r_e} - 2 \|\overrightarrow{\mathbf{OJ}}\|\mathbf{l}_i,
    \label{equa:r_e}
\end{align}
where $w$ and $h$ are the width and height of the $I_i$ respectively,  $\|\overrightarrow{\mathbf{OJ}}\|$ is computed via
\begin{equation}
    \|\overrightarrow{\mathbf{OJ}}\| = -\frac{\|\overrightarrow{\mathbf{OH}}\|}{ \langle \mathbf{l}_i, \mathbf{s}_i \rangle} = -\frac{d}{\langle \mathbf{l}_i, \mathbf{s}_i \rangle},
    \label{equa:OJ}
\end{equation}
and $\mathbf{r}$ and $\mathbf{c}$ are two orthonormal vectors parallel to the projection screen and determined by
\begin{equation}
    \mathbf{c} = \begin{cases}
\frac{(-\mathbf{s}_i[1], \mathbf{s}_i[0], 0)}{\left\| (-\mathbf{s}_i[1], \mathbf{s}_i[0], 0) \right\|
} & \text{if } \mathbf{s}_i \neq (0, 0, 1) \\
(0, 1, 0) & \text{if } \mathbf{s}_i = (0, 0, 1)
\end{cases},
\qquad \quad
    \mathbf{r} = \mathbf{c} \times \mathbf{s}_i.
    \label{equa:r}
\end{equation}
Each ray is then discretized into points by sampling randomly within $n$ equally spaced segments.
The occupancy value of the entire ray is determined by the pixel value of $(p_x, p_y)$ in the target image: if $(p_x,p_y)$ is black, then the ray occupancy value is set to 1; otherwise, it is set to 0.
We traverse all pixels of the input image, generating the corresponding rays and occupancy values to construct the "Ray-Occupancy" dataset.

\paragraph*{Rendering Loss.} 
To enforce matching between the geometry's projections and the target images, we introduce a rendering loss $\mathcal{L}_{\mathrm{ren}}$. For a batch \(\mathcal{B}\) of rays, it is defined as:
\begin{equation}
    \mathcal{L}_{\mathrm{ren}} = \frac{\alpha}{|\mathcal{B}|}\sum\nolimits_{j=1}^{|\mathcal{B}|}(M_{j}-O_{j})^2.
    \label{equa:Lren}
\end{equation}
Here, $M_j$ is the ground-truth occupancy label for the $j$-th ray in the batch. The weighting factor $\alpha$ is the maximum ratio of input image area to the area of the shadow region's bounding box in the image; this weighting gives greater importance to the rendering loss in cases of smaller shadows within the target images, ensuring they are not overlooked during training.
The term $O_{j}$ is the predicted ray occupancy, computed by sampling the implicit function along rays.

\begin{wrapfigure}{r}{0.25\linewidth}
    \centering
    \vspace{-6pt}
    \includegraphics[trim=0 0 0 0,clip,width=1.0\linewidth]{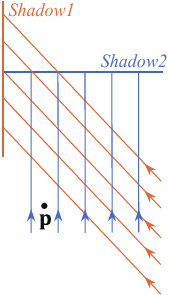}
    \vspace*{-12pt}
\end{wrapfigure}
\begin{figure}[t]
\centering
\includegraphics[width=0.9\columnwidth]{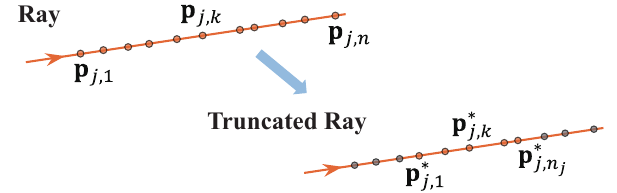}
\caption{\textbf{Visualization of the Quantities in Eq. (\ref{equa:O_ij}).} We demonstrate the truncation of a given ray, with the truncated points shown in gray.} 
\label{fig:Truncated Ray visual}
\end{figure}

An important consideration for this computation is ray truncation. When dealing with multiple shadow constraints, a sampling point $\mathbf{p}$ along a ray defined for one constraint (e.g., Shadow2) might lie 
outside the viewing frustum of another (e.g., Shadow1), as depicted conceptually from a top-down view in the inset figure. If $\mathbf{p}$ were occupied, it could cast an erroneous shadow outside the intended region for Shadow1. To resolve this, we truncate all rays to the intersection of the viewing frustums generated by all projection constraints. This truncation choice is also applied during the final reconstruction phase explained in Sec.~\ref{sec_rec}, to ensure the final result does not introduce unnecessary projections. Additionally, using input images with a wide white margin can also help mitigate these adverse effects.
The predicted occupancy $O_{j}$ is then computed using points from these truncated rays via:
\begin{equation}
    O_{j} = 1 - \prod\nolimits_{k =1}^{n_{j}} (1 - f_{\Theta,\mathbf{l},\mathbf{s}}(\mathbf{p}^*_{j,k})),
    \label{equa:O_ij}
\end{equation}
where \(\{\mathbf{p}^*_{j,k}\}_{k=1}^{{n}_{j}}\) is the set of points on the truncated ray $j$, and \({n}_{j}\) is the number of the sampling points on the truncated ray (see Fig.~\ref{fig:Truncated Ray visual}).

Similar to~\cite{liu2020general}, this formulation ensures that $O_j$
approaches $1$ if any point on the ray is occupied, providing the flexibility needed to satisfy multiple constraints simultaneously. Furthermore, because each point holds the same significance, this expression is unbiased.

\paragraph*{Cohesion Loss.} 
The requirement for a ray to be occupied can be satisfied by many different combinations of occupancy values at the sample points along that ray. For instance, a single occupied point or several disconnected occupied points would both satisfy the constraint. To produce solid shapes that are easy to fabricate, we must encourage the occupied points (where occupancy is close to 1) to be concentrated together. Therefore, we introduce a cohesion loss $\mathcal{L}_{\mathrm{coh}}$ that penalizes changes in occupancy between adjacent points along a truncated ray, to discourage multiple thin layers and promote contiguous segments of occupied space:
\begin{equation}
    \mathcal{L}_{\mathrm{coh}} = \frac{1}{|\mathcal{B}|}\sum\nolimits_{j=1}^{|\mathcal{B}|} \frac{1}{n_j}\sum\nolimits_{k=1}^{n_j-1} (f_{\Theta,\mathbf{l},\mathbf{s}}(\mathbf{p}^*_{j,k+1}) - f_{\Theta,\mathbf{l},\mathbf{s}}(\mathbf{p}^*_{j,k}))^2. \label{geo_loss1}
\end{equation}

\paragraph*{Smoothness Loss.} 
To produce visually appealing geometries that are suited for physical fabrication, we also introduce a loss to enforce smoothness of the resulting surface.
To this end, we must first identify points that lie on the implicit surface.
We note that by definition, the magnitude of the occupancy function's gradient \(\|\nabla f_{\Theta,\mathbf{l},\mathbf{s}}\|\) should be significantly larger at the surface boundary than inside or outside the geometry. Utilizing this property, we look for points $\{\hat{\mathbf{p}}\}$ that satisfy the following criteria as the surface points:
\begin{equation}
    \|\nabla f_{\Theta,\mathbf{l},\mathbf{s}}(\hat{\mathbf{p}})\| > \theta w,
    \label{equa:judge}
\end{equation}
where $w$ is the width of the input image, and \(\theta\) is a tunable threshold. According to our sampling strategy, the spacing between sampling points in 3D space is approximately on the order of 1/$w$ where $w$ is the image width. Since the occupancy value typically transitions from 1 to 0 across the surface boundary, the estimated surface gradient is roughly on the order of 1/(1/$w$) = $w$. We additionally introduce a scaling parameter to adjust this estimation as needed.

Estimating the occupancy gradient via direct network backpropagation can be unstable, as the network may exhibit sharp changes at scales smaller than our sampling resolution, leading to erroneously large gradient estimates. 
Instead, we use the first-order finite difference to estimate \(\nabla f_{\Theta, \mathbf{l}, \mathbf{s}}(\hat{\mathbf{p}})\), which should satisfy:
\begin{equation}
    \nabla f_{\Theta, \mathbf{l}, \mathbf{s}}(\hat{\mathbf{p}}) \cdot (\mathbf{p}' - \hat{\mathbf{p}}) = f_{\Theta, \mathbf{l}, \mathbf{s}}(\mathbf{p}') - f_{\Theta, \mathbf{l}, \mathbf{s}}(\hat{\mathbf{p}}),
    \label{equa:directional gradient}
\end{equation}
where $\mathbf{p}'$ is a neighboring point. Then by collating such conditions for $k_1$ nearest points $\{\mathbf{p}'\}$ for $\hat{\mathbf{p}}$, we obtain an over-determined linear system for \(\nabla f_{\Theta, \mathbf{l}, \mathbf{s}}(\hat{\mathbf{p}})\):
\begin{equation}
    \mathbf{K} \nabla f_{\Theta, \mathbf{l}, \mathbf{s}}(\hat{\mathbf{p}}) = \mathbf{b},
    \label{equa:matrix}
\end{equation}
where the matrix \(\mathbf{K} \in \mathbb{R}^{k_1 \times 3}\) stores the vectors $\{\mathbf{p}' - \hat{\mathbf{p}}\}$ in its rows, and the vector \(\mathbf{b} \in \mathbb{R}^{k_1}\) contains the values $\{f_{\Theta, \mathbf{l}, \mathbf{s}}(\mathbf{p}') - f_{\Theta, \mathbf{l}, \mathbf{s}}(\hat{\mathbf{p}})\}$ in its components. We then solve this system in a least-squares way to obtain $\nabla f_{\Theta, \mathbf{l}, \mathbf{s}}(\hat{\mathbf{p}})$. We perform this procedure for all sample points in the current batch of truncated rays, and identify the set $\widehat{\mathcal{P}}$ of surface points among them according to condition~\eqref{equa:judge}. 

\begin{figure}[t]
\centering
\includegraphics[width=\columnwidth]{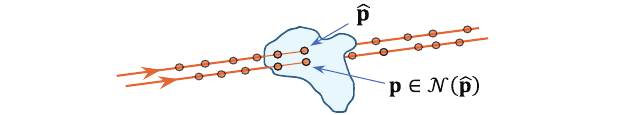}
\caption{\textbf{Visualization of the Quantities in Eq. (\ref{geo_loss2}).} We demonstrate how to find \(k_2\) nearest points for the point $\hat{\mathbf{p}}$ on the geometry surface.} 
\label{fig:Intersect Visual}
\end{figure}

With the surface points identified, we further note that the gradient $\nabla f_{\Theta, \mathbf{l}, \mathbf{s}}(\hat{\mathbf{p}})$ should align with the surface normal at $\hat{\mathbf{p}}$ up to a roughly constant scaling factor.
Therefore, we introduce the following loss term that enforces surface smoothness by penalizing the rate of normal changes between adjacent surface points:
\begin{equation}
    \mathcal{L}_{\mathrm{smo}} = \frac{1}{|\widehat{\mathcal{P}}|}
    \sum\nolimits_{\hat{\mathbf{p}} \in \widehat{\mathcal{P}}} 
    \frac{1}{|\mathcal{N}(\hat{\mathbf{{p}}})|}\sum\nolimits_{\mathbf{p}\in \mathcal{N}(\hat{\mathbf{{p}}})}\frac{\|\nabla f_{\Theta, \mathbf{l}, \mathbf{s}}(\hat{\mathbf{{p}}})-\nabla f_{\Theta, \mathbf{l}, \mathbf{s}}(\mathbf{p})\|}{\left\| \hat{\mathbf{{p}}} - \mathbf{p} \right\|},
    \label{geo_loss2}
\end{equation}
where $\mathcal{N}(\hat{\mathbf{{p}}})$ denotes the set of $k_2$ nearest surface points for $\hat{\mathbf{{p}}}$ (see Fig.~\ref{fig:Intersect Visual}).

\paragraph*{Volume Loss.} 
To encourage material efficiency and enhance artistic appeal, we introduce a volume loss $\mathcal{L}_{\mathrm{vol}}$. Minimizing the volume directly addresses industrial requirements for lower material usage. Aesthetically, this is crucial for avoiding trivial solutions such as the intersection of simple cylinders extruded along light directions, and instead discovering more intricate, non-obvious shapes. We formulate this as a differentiable approximation of the object's volume:
\begin{equation}
    \mathcal{L}_{\mathrm{vol}} = \frac{1}{|\mathcal{B}|} \sum\nolimits_{j=1}^{|\mathcal{B}|} \sum\nolimits_{k=1}^{n_{j}} \frac{\omega_{j,k}}{1 + \exp\left(-\frac{f_{\Theta,\mathbf{l},\mathbf{s}}(\mathbf{p}_{j,k}^*) - \tau}{T}\right)}, \label{vol_loss} 
\end{equation}
where \( \omega_{j,k} \) is a quantity proportional to the volume associated with a sample point $\mathbf{p}_{j,k}^*$ on the truncated ray $j$:
\begin{equation}
\omega_{j,k} = 
\begin{cases} 
\|\mathbf{p}_{j,2}^* - \mathbf{p}_{j,1}^*\| & \text{if } k = 1 \\ 
\|\mathbf{p}_{j,n_{j}}^* - \mathbf{p}_{j,n_{j}-1}^*\| & \text{if } k = n_{j} \\ 
\frac{1}{2} \|\mathbf{p}_{j,k}^* - \mathbf{p}_{j,k-1}^*\| + \frac{1}{2} \|\mathbf{p}_{j,k+1}^* - \mathbf{p}_{j,k}^*\| & \text{otherwise}.
\end{cases}
\end{equation}
The factor $(1 + \exp(-\frac{f_{\Theta,\mathbf{l},\mathbf{s}}(\mathbf{p}_{j,k}^*) - \tau}{T}))^{-1}$ in Eq.~(\ref{vol_loss}) is a sigmoid function that acts as a differentiable "soft switch" that activates based on the occupancy value of point $\mathbf{p}_{j,k}^*$  relative to a threshold $\tau$, while \( T \) is a temperature parameter controlling the softness of the transition. 
This soft counting strategy, when summed over all points, yields a differentiable approximation of the total volume.

\paragraph*{Binarization Loss.} 
To ensure the learned implicit field represents a solid, well-defined object, the occupancy values should converge to either 0 or 1.
This prevents the network from using intermediate values (i.e., transparency) as a way to minimize the rendering loss without forming a clear surface. 
To enforce this, we apply a binarization loss:
\begin{equation}
    \mathcal{L}_{\mathrm{bin}} = \frac{1}{|\mathcal{B}|} \sum\nolimits_{j=1}^{|\mathcal{B}|} \frac{1}{n_j} \sum\nolimits_{k=1}^{n_j} \min \left( f_{\Theta,\mathbf{l},\mathbf{s}}^2(\mathbf{p}^*_{j,k}), \left( 1 - f_{\Theta,\mathbf{l},\mathbf{s}}(\mathbf{p}^*_{j,k}) \right)^2 \right). \label{regular_loss}
\end{equation}
This loss penalizes any predicted occupancy value that is not close to either 0 or 1, pushing the network's outputs toward a binary state.

\paragraph*{Total Loss.} 
The total loss for a training batch is a weighted sum of the components described above:
\begin{equation}
    \mathcal{L} = \mathcal{L}_{\mathrm{ren}} + \beta_1\mathcal{L}_{\mathrm{coh}} + \beta_2\mathcal{L}_{\mathrm{smo}} +\beta_3\mathcal{L}_{\mathrm{vol}}+\beta_4\mathcal{L}_{\mathrm{bin}},
    \label{eq:FinalLoss}
\end{equation}
where \( \beta_1 \), \( \beta_2 \), \( \beta_3 \), and \( \beta_4 \) are adjustable hyperparameters that balance the influence of each loss component. 

Our framework is end-to-end differentiable, which allows us to jointly train all learnable parameters including the network weights $\Theta$, the light directions $\{\mathbf{l}_i\}$, and the screen orientations $\{\mathbf{s}_i\}$.
The training process iterates through batches of rays from the dataset, and update all parameters simultaneously via backpropagation. This unified training process allows the geometry and light conditions to be refined together with every batch, iteratively improving the match to the shadow constraints.

\subsection{Registration} \label{training}

To better evaluate the consistency between the rendered shadows and the target images, our framework accommodates rigid transformations of the projection targets. This process is performed periodically after every five epochs, as a single epoch may not produce sufficient refinement for a meaningful registration.

To do so, we first render the current shadow projections from the latest learned geometry. Next, we extract the boundary point clouds from both the rendered shadows and the target shadows. We then use the Iterative Closest Point (ICP) algorithm~\cite{Besl1992} to register the target shadow boundary point cloud to the rendered shadow boundary point cloud. The resulting rigid transformation is then applied to the target shadow to derive an updated target image that maintains the target shadow shape while being closer to the current rendered shadow. In this way, we allow the model to converge on a solution that is consistent with the input shadows up to rigid motions.

\subsection{Reconstruction}\label{sec_rec}

Once the iterative training and registration process is complete, the final output is the learned neural implicit occupancy function $f_{\Theta,\mathbf{l},\mathbf{s}}$. From this implicit field, we extract an explicit mesh surface suitable for simulation and physical fabrication. This is achieved by first evaluating the function on a high-resolution 3D grid. We then apply the Marching Cubes algorithm~\cite{WH87} to this grid to extract the isosurface corresponding to a specific occupancy threshold $\tau$ (we use $0.5$ in our experiments), obtaining the final 3D mesh.
\section{Experiments}\label{sec:exp}

\paragraph*{Implementation Details.}
For the occupancy function described in Sec.~\ref{subsec:occu_net}, input coordinates \(\left(x, y, z\right)\) are normalized to the range \([-0.5, 0.5]\). The positional encoding in Eq.~(\ref{equa:gamma_p}) uses $L=6$. The MLP \( \widetilde{f}_{\Theta, \mathbf{l}, \mathbf{s}} \) in Eq.~(\ref{eq:MLPFunc}) consists of 8 fully connected layers, each with 256 channels and ReLU activation functions, except for the final layer which uses a sigmoid activation to output values in  \([0, 1]\). 
When generating the ``Ray-Occupancy'' dataset in Sec. \ref{one_epoch_training}, the distance $d$ from the projection plane to the origin is set to $0.5$,  consistent with the normalized coordinate range. The number of sampling points along each ray is set to $n = w$. The truncated region is defined by the intersection of shadow constraints, which are represented as prism-like volumes obtained by translating the input image parallel to the light direction. 
For identifying surface points in Eq.~(\ref{equa:judge}), we use $\theta=0.4$.
The gradient estimation in Eq.~(\ref{equa:matrix}) employs $k_1=26$ points, and the smoothness loss in Eq.~(\ref{geo_loss2}) uses  $k_2=6$ nearest neighbors.  
The loss weights in Eq.~(\ref{eq:FinalLoss}) are scheduled during training as follows: \( \beta_1 = 10^{-3} \times 2^{\min(\text{epoch}, 3)} \); $\beta_2$ and $\beta_3$ are set to $10^{-4}$ if $\text{epoch} > 3$ and $0$ otherwise; and \( \beta_4 = 5 \times 10^{-2} \times 2^{\min(\text{epoch}, 3)}\). 
The weights \( \beta_1 \) for cohesion and \( \beta_4 \) for binarization are increased early to accelerate model convergence. The smoothness loss $\mathcal{L}_{\mathrm{smo}}$ and the volume loss $\mathcal{L}_{\mathrm{vol}}$ are only activated via $\beta_2$ and $\beta_3$ after epoch 3, allowing the basic shape to form first. 
We implement our method using Pytorch~\cite{Pytorch}, and use the Adam optimizer for 30 epochs. For surface extraction via Marching Cubes (Sec.~\ref{sec_rec}),
we sample the occupancy field on a 200$^3$ grid.
All experiments were performed on an NVIDIA RTX 3090 GPU with 24GB VRAM. It typically takes around 1.5 hours to obtain a stable geometry.

\paragraph*{Qualitative Comparison.} We begin by summarizing the advantages of our framework over previous methods~\cite{MP09,STR22} in Tab.~\ref{tab:advantage}. A key aspect is handling potentially incompatible input constraints. While \cite{MP09} applies an As-Rigid-As-Possible (ARAP) transformation, this often leads to significant projection distortions; meanwhile, \cite{STR22} does not explicitly address such compatibility issues. 
Prior works also lack flexibility in light and screen setups; they typically do not test complex light conditions or support optimization of light directions and screen orientations.
For complex topologies, \cite{MP09} handles certain cases partially but may introduce distortions due to large ARAP deformations. In contrast, the voxel-based approach in \cite{STR22} is limited by resolution and can produce floating regions or non-watertight meshes, while their mesh-based approach struggles with complex topologies, often causing mesh flips as it deforms an initial sphere without changing topology. 
Finally, volume optimization, which is crucial for fabrication and artistic aesthetics (e.g., avoiding trivial cylindrical intersections), is overlooked in these previous works.
\begin{table}[t]
\centering
\caption{\textbf{Advantages of Our Model Compared with Previous Works.} 
We demonstrate the superiority of our method in four aspects. 
$\times$ indicates that the consideration is either not made or the result is insufficient, 
while $\checkmark$ signifies that the consideration is taken into account and a good result is achieved.}
\resizebox{\linewidth}{!}{%
\begin{tabular}{c|c|c|c|c}
\hline
\multirow{2}{*}{} 
& \multirow{2}{*}{\cite{MP09}} 
& \multicolumn{2}{c|}{\cite{STR22}} 
& \multirow{2}{*}{Ours} \\
\cline{3-4}
& & Voxel-based & Mesh-based & \\
\hline
\multirow{2}{*}{\textbf{\shortstack{Incompatible \\ Inputs}}} 
& \multirow{2}{*}{\(\checkmark\)} 
& \multirow{2}{*}{\(\times\)} 
& \multirow{2}{*}{\(\times\)} 
& \multirow{2}{*}{\textbf{\(\checkmark\)}} \\
& & & & \\
\hline
\multirow{2}{*}{\textbf{\shortstack{Light and Screen \\ Flexibility}}} 
& \multirow{2}{*}{\(\times\)} 
& \multirow{2}{*}{\(\times\)} 
& \multirow{2}{*}{\(\times\)} 
& \multirow{2}{*}{\textbf{\(\checkmark\)}} \\
& & & & \\
\hline
\multirow{2}{*}{\textbf{\shortstack{Complex \\ Topology}}} 
& \multirow{2}{*}{\(\times\)} 
& \multirow{2}{*}{\(\checkmark\)} 
& \multirow{2}{*}{\(\times\)} 
& \multirow{2}{*}{\textbf{\(\checkmark\)}} \\
& & & & \\
\hline
\multirow{2}{*}{\textbf{\shortstack{Volume \\ Optimization}}} 
& \multirow{2}{*}{\(\times\)} 
& \multirow{2}{*}{\(\times\)} 
& \multirow{2}{*}{\(\times\)} 
& \multirow{2}{*}{\textbf{\(\checkmark\)}} \\
& & & & \\
\hline
\end{tabular}%
}
\label{tab:advantage}
\end{table}

\begin{figure*}[t]
\centering
\includegraphics[width=0.98\textwidth]{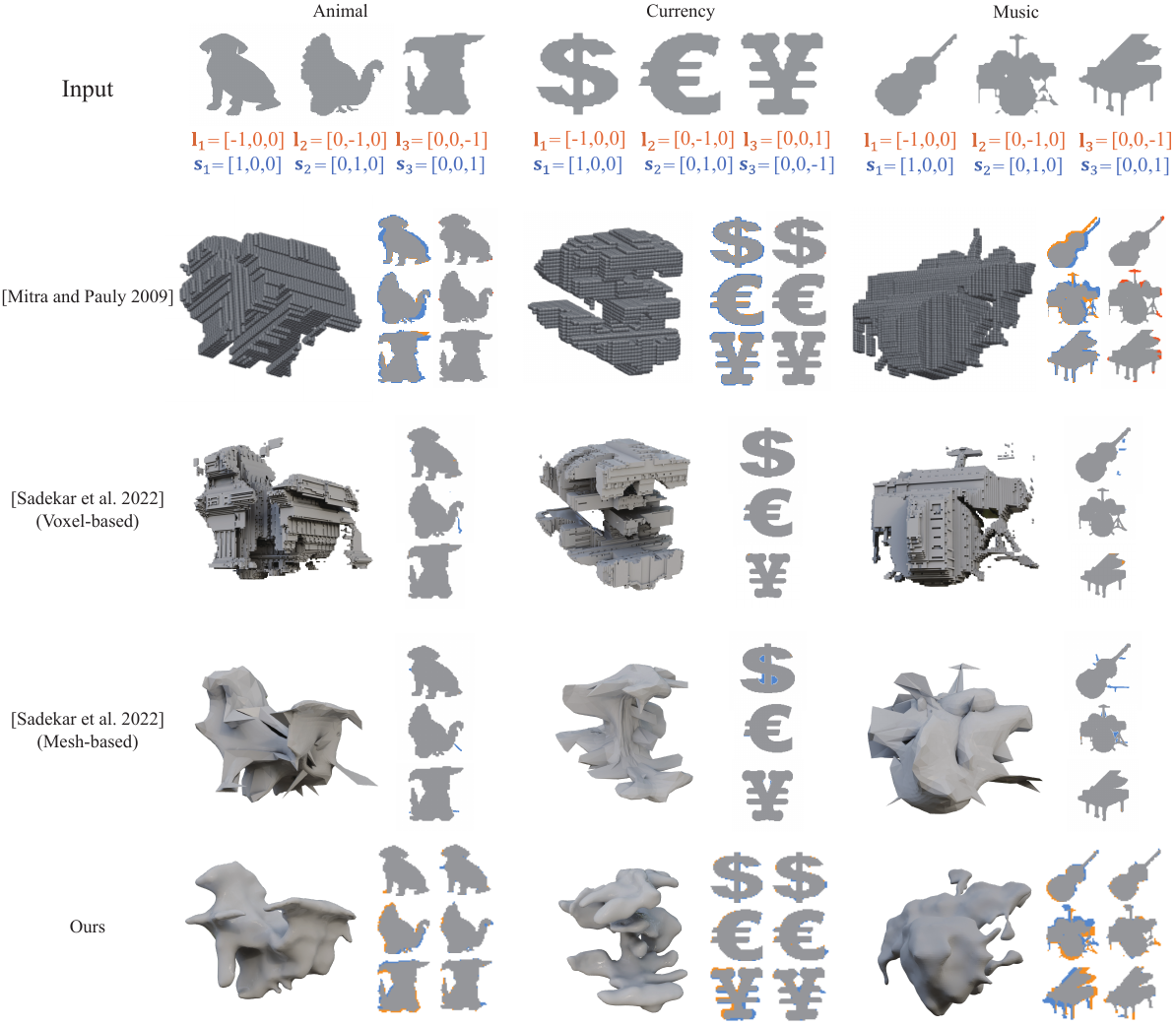}
\caption{\textbf{Comparison with Previous Works.} For \cite{MP09}, the first column shows inputs vs. deformed target projections, and the second shows deformed target projections vs. final projections. For \cite{STR22}, the column represents inputs vs. final projections without deformation. For ours, the first column shows inputs vs. rigid registrations, and the second shows rigid registrations vs. final projections. } \label{fig:tests}
\end{figure*}

Our primary comparisons are made against previous works specifically targeting shadow art \cite{MP09,STR22}. We did not include methods such as NeuS~\cite{WLL*21} or NeuralUDF~\cite{LLL*23} as baselines because, to the best of our knowledge, there are currently no shadow art approaches that adopt SDF or UDF as representations. Fig.~\ref{fig:tests} visually compares our approach with \cite{MP09,STR22}. Both methods are run using the implementations provided by the authors\footnote{\url{https://graphics.stanford.edu/~niloy/research/shadowArt/shadowArt_sigA_09.html}}\footnote{\url{https://github.com/kaustubh-sadekar/ShadowArt-Revisited}}. Our method produces more robust and coherent results with smoother geometry compared to these prior methods. In contrast, the voxel-based methods from~\cite{MP09,STR22} suffer from discretization artifacts and yield non-smooth structures, while the mesh-based method from~\cite{STR22} may produce incoherent surfaces prone to flipped triangles.  Moreover, our results are notably more compact thanks to the volume loss. Unlike previous approaches where the results resemble intersecting geometric cylinders, our results more effectively capture the artistic intentions of shadow art, producing visually striking effects.

\begin{figure*}[t]
\centering
\includegraphics[width=\textwidth]{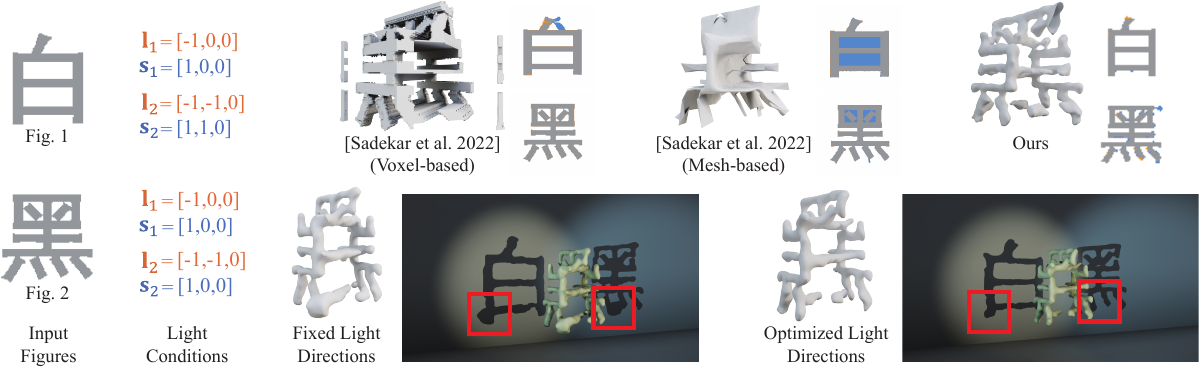}
\caption{\textbf{Experiments under Various Light Conditions.} (First line:) We present results under fixed light conditions with the light direction perpendicular to the projection plane, and compare them with \cite{STR22}, showing that our model achieves superior results even without light condition optimization. (Second line:) We also demonstrate the impact of different light directions on projection accuracy, proving that optimized light directions significantly improve the projection results.} \label{fig:abl_angle}
\end{figure*}

For inputs with complex topology, Fig.~\ref{fig:abl_angle} further highlights the difference between our method and \cite{STR22}. The voxel-based method in \cite{STR22} can fail to produce the correct geometry and may generate structures extending beyond the renderer's viewable range, causing extraneous projections. This shows the importance of the ray truncation discussed in Sec.~\ref{one_epoch_training}. The mesh-based method in \cite{STR22} completely fails with such complex topologies, whereas our method yields accurate and geometrically stable results. Fig.~\ref{fig:abl_angle} also demonstrates the importance of optimizing light directions. With fixed light directions, projection accuracy can be severely compromised (e.g., the illegible characters in the red-boxed region). Optimizing light directions significantly enhances projection accuracy.

\paragraph*{Quantitative Comparison.}
To measure projection accuracy, we compute the Intersection over Union (IoU)~\cite{IOU} and Dice Similarity (DS)~\cite{DS} between our final projected shadows and the (rigidly registered) input target images. As shown in Tab.~\ref{tab:accuracy}, when compared against \cite{MP09} and both the voxel-based and mesh-based approaches of \cite{STR22}, our method achieves the highest scores on both metrics across various test cases.

To demonstrate the effectiveness of our volume optimization, we compare the surface area and volume of our generated meshes against those from the voxel-based method in \cite{STR22} (the mesh-based method in \cite{STR22} often fails to produce manufacturable meshes for complex cases). Tab.~\ref{tab:volume} shows that our approach yields significant reductions in both surface area and volume, highlighting improved material efficiency.

\begin{figure}[t]
\centering
\includegraphics[width=1.0\columnwidth]{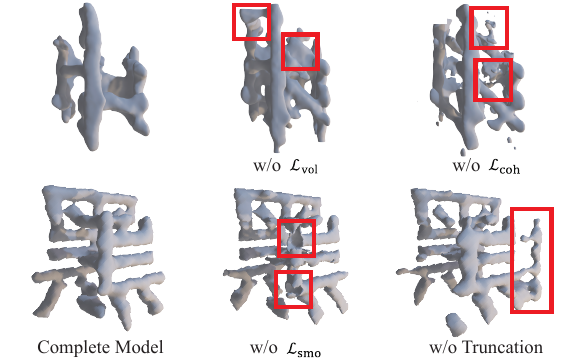}
\caption{\textbf{Ablation Studies of Losses and Truncation.} We omitted the cohesion, smoothness, volume loss terms and the truncation technique respectively, while keeping all other parameters the same. Red boxes highlight unreasonable geometric structures.} \label{fig:abl_loss}
\end{figure}

\paragraph*{Ablation Studies.} 
We first conducted an ablation study to verify the necessity of our loss functions and truncation technique for achieving high-quality and reasonable geometry. In this experiment, when the ablation target is a loss term, the corresponding parameter is set to zero; when the ablation target is the truncation operation, it is omitted. All other operations in Sec.~\ref{sec-method} remain the same, with the same parameters as those in Sec.~\ref{sec:exp}. As illustrated in Fig.~\ref{fig:abl_loss}, removing the volume loss $\mathcal{L}_{\mathrm{vol}}$ results in significantly redundant geometric structures, while omitting the cohesion loss $\mathcal{L}_{\mathrm{coh}}$ or the smoothness loss $\mathcal{L}_{\mathrm{smo}}$ leads to multi-layered structures and local inconsistencies, respectively, severely degrading the geometric quality. Without the truncation operation, large structures extending beyond the renderer's viewable range greatly impair the projection results. Unreasonable geometric structures are highlighted with red boxes. Furthermore, without the binarization loss $\mathcal{L}_{\mathrm{bin}}$, the model fails to train effectively.

\begin{figure}[t!]
\centering
\includegraphics[width=1.0\columnwidth]{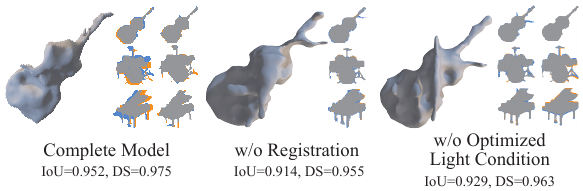}
\caption{\textbf{Ablation Studies of Registration and Optimized Light Condition.} We present the results with the registration operation and the optimization of light condition (light directions and screen orientations) omitted respectively, while keeping all parameters the same. IoU and DS values are averaged across all images.}\label{fig:abl_teq}
\end{figure}

We also conducted an ablation study to demonstrate the effectiveness of our proposed registration operation and the simultaneous optimization of light directions and screen orientations on projection accuracy. In this experiment, all parameters are kept the same as in Sec.~\ref{sec:exp}, while either the registration or the light condition optimization is omitted in the corresponding ablations. As shown in Fig.~\ref{fig:abl_teq}, the geometric results, along with the IoU and DS metrics, clearly indicate that removing either component leads to a decrease in the accuracy of the final projection.

\begin{table}[t]
\centering
\caption{\textbf{Comparison of Accuracy with the (Rigidly Registered) Input Images.} Our comparison focuses on the alignment between the projections of the output geometries and the (rigid registered) input projections. The values are averaged across all images. A higher value indicates closer resemblance to the input images.}
\setlength{\tabcolsep}{2.5pt} 
\begin{tabular}{c|c|cc|cc|cc}
\hline
\multicolumn{2}{c|}{\multirow{2}{*}{\textbf{Method}}}
 & \multicolumn{2}{c|}{\textbf{Animal}} 
 & \multicolumn{2}{c|}{\textbf{Currency}} 
 & \multicolumn{2}{c}{\textbf{Music}} \\
 \multicolumn{2}{c|}{}  & \textbf{IoU} & \textbf{DS} & \textbf{IoU} & \textbf{DS} & \textbf{IoU} & \textbf{DS} \\
\hline
\multicolumn{2}{c|}{\cite{MP09}} & 0.823 & 0.903 & 0.867 & 0.929 & 0.750 & 0.853 \\
\hline
\multirow{2}{*}{\cite{STR22}} & Voxel-based & 0.876 & 0.934 & 0.916 & 0.956 & 0.901 & 0.948 \\ \cline{2-8}
& Mesh-based  & 0.847 & 0.917 & 0.870 & 0.930 & 0.853 & 0.920 \\
\hline
\multicolumn{2}{c|}{Ours} &  \textbf{0.967} & \textbf{0.983} & \textbf{0.961} & \textbf{0.980} & \textbf{0.952} & \textbf{0.975}\\
\hline
\end{tabular}
\label{tab:accuracy}
\end{table}

\begin{table}[t]
\centering
\caption{\textbf{Comparison of Surface Area and Volume of Final Mesh.} Area and Vol. represent the surface area and volume computed using MeshLab, with the bounding box diameter normalized to 1. The unit of volume is $10^{-2}$.}
\setlength{\tabcolsep}{3pt} 
\begin{tabular}{c|cc|cc|cc}
\hline
\multirow{2}{*}{\textbf{Method}} & \multicolumn{2}{c|}{\textbf{Animal}} & \multicolumn{2}{c|}{\textbf{Currency}} & \multicolumn{2}{c}{\textbf{Music}} \\
 & \textbf{Area} & \textbf{Vol.} & \textbf{Area} & \textbf{Vol.} & \textbf{Area} & \textbf{Vol.} \\
\hline
 \cite{STR22} (Voxel-based) & 1.57 & 4.36 & 2.02 & 3.63 & 1.05 & 2.61 \\
\hline
Ours & \textbf{1.03} & \textbf{3.35} & \textbf{1.04} & \textbf{2.13} & \textbf{0.76} & \textbf{1.93} \\
\hline
\end{tabular}
\label{tab:volume}
\end{table}

\begin{figure*}[!htb]
\centering
\includegraphics[width=1.0\textwidth]{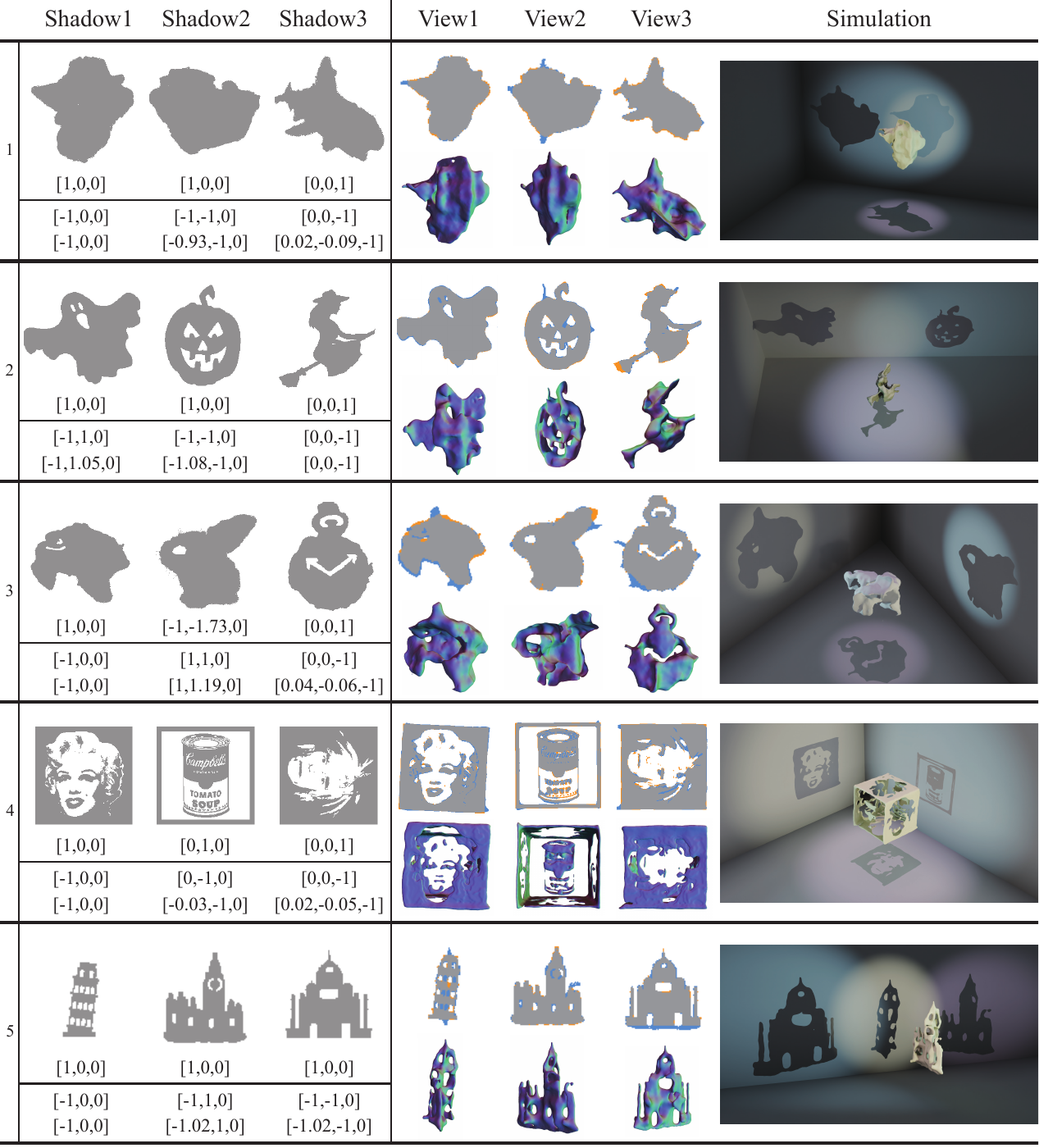}
\caption{\textbf{Gallery of Neural Shadow Art}. We present the input images, the corresponding initial (second-line number) and optimized (third-line number) lighting directions and screen orientations (first-line number), the differentiable rendering results and the normals of the reconstructed geometry, as well as the simulation results conducted in Blender. In the differentiable rendering results, blue and orange represent the parts unique to our results and the input, respectively, while gray represents the shared parts.} \label{fig:main_results}
\end{figure*}
\begin{figure}[t]
\centering
\includegraphics[width=1.\columnwidth]{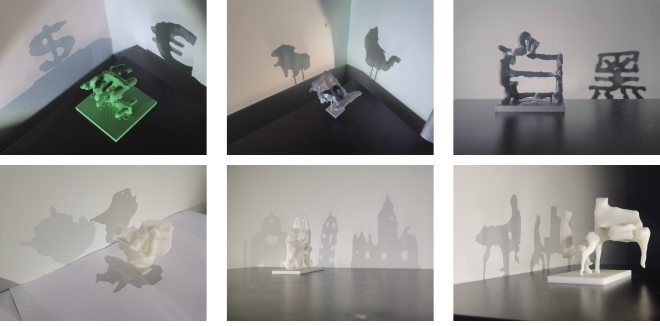}
\caption{\textbf{Fabrication Results.} Results show that the manufactured outcomes produce shadow effects nearly identical to the simulation, delivering stunning artistic effects.} \label{fig:fabrication}
\end{figure}

\paragraph*{Gallery and Fabrication.}
We demonstrate the capabilities of our Neural Shadow Art framework on a diverse set of inputs, varying in image topology and lighting complexity. Fig.~\ref{fig:main_results} presents a gallery of results. For these examples, we showcase outcomes where screen orientations are fixed while light directions are optimized, highlighting the framework's flexibility in handling varied light conditions. The generated meshes exhibit reasonable geometry, and the rendered projections closely align with the target input images, even for intricate topologies and complex, optimized lighting systems—a capability not demonstrated in previous works.

To validate the manufacturability of our designs, we 3D-printed several of the generated geometries. As shown in Fig.~\ref{fig:fabrication}, the fabricated objects successfully produce the intended shadow effects, closely matching the simulations and delivering compelling artistic results.

\section{Conclusion $\&$ Discussion}

In this work, we presented Neural Shadow Art, a novel approach utilizing a neural implicit occupancy representation, significantly expanding the capabilities of computational shadow art design. Our method overcomes key limitations of prior art, offering freedom from the fixed resolution constraints of voxel-based methods and the predefined topology restrictions of mesh-based techniques. Neural Shadow Art can handle complex lighting conditions, including scenarios where light is not perpendicular to the projection plane, and supports the joint optimization of these lighting parameters. Consequently, our approach generates 3D geometries that achieve superior alignment with target projection constraints, especially for inputs with complex topologies, while simultaneously promoting lower material usage and maintaining excellent geometric properties suitable for fabrication. These capabilities are important for both efficient industrial production and for enhancing the artistic appeal of shadow art.

Despite these advancements, our method has certain limitations. Currently, it requires precise 2D binary images as projection targets and lacks support for more abstract or creative input modalities, such as semantically guided design, which would be a valuable extension for artists. Additionally, while our experiments show a high success rate for fabricating shadow art with up to three constraints, designs involving four or more constraints often require careful adaptation of the input images to achieve satisfactory results due to the excessive number of constraints. Another challenge arises with disconnected input silhouettes: the resulting 3D geometry may also consist of disconnected components, posing difficulties for 3D printing due to floating parts. We currently mitigate this by either embedding the geometry within a supporting volume of translucent material or by manually adding small connecting rods. A promising direction for future work is to employ heuristic learning methods—such as correcting connectivity across multi-directional projections—or to explicitly enhance connectivity within the neural representation itself, in order to ensure manufacturable single-piece outputs.

\section*{Acknowledgements}
This research was supported by the National Natural Science Foundation of China(No.62441224, No.62272433), the Fundamental Research Funds for the Central Universities, and the USTC Fellowship (No.S19582024, No.U19582024).

\bibliographystyle{eg-alpha-doi} 
\bibliography{egbibsample}

\end{document}